\documentclass[a4paper,fleqn]{cas-dc}

\usepackage[authoryear,longnamesfirst]{natbib}
\usepackage{amsmath}
\usepackage{graphicx}
\usepackage{subcaption}

\def\tsc#1{\csdef{#1}{\textsc{\lowercase{#1}}\xspace}}
\tsc{GEI}
\tsc{CNN}

\begin{document}
\let\WriteBookmarks\relax
\def\floatpagepagefraction{1}
\def\textpagefraction{.001}

\shorttitle{Two-Stage Multi-View Gait Recognition}

\shortauthors{Long Hoang Le, Trung Thanh Ngo}

\title [mode = title]{Two-Stage Multi-View Gait Recognition with a Re-Embedding Network}

\tnotemark[1]

\tnotetext[1]{}

%

\author[1]{Long Hoang Le}

\cormark[1]

\fnmark[1]

\ead{hoanglong1712@gmail.com}

\ead[url]{}

\credit{Conceptualization, Methodology, Software, Validation, Writing -- original draft}

\author[1]{Trung Thanh Ngo}

\cormark[2]

\fnmark[2]

\ead{trungnth@soict.hust.edu.vn}

\ead[url]{}

\credit{ Supervisor -- original draft}

\affiliation[1]{organization={Hanoi University of Science and Technology},
            addressline={Dai Co Viet},
            city={Hanoi},
            postcode={100000},
            country={Vietnam}}

\cortext[1]{Corresponding author}

\fntext[1]{}


\begin{abstract}
Gait recognition always remains challenging due to severe overfitting and the rigid view constraints common in single-stage approaches. We propose a two-stage framework, termed Translate-First-Then-Reason (TFTR), to address these issues. In the first stage, a shallow Siamese convolutional network with triplet loss maps Gait Energy Images (GEIs) into a 128-dimensional view-specific embedding space. In the second stage, these per-view embeddings are treated as tokens and processed by a 12-layer Transformer encoder, which re-projects them into a new space with improved cosine separability. This design enables flexible fusion of an arbitrary number of views at inference, overcoming the fixed-input limitations of prior methods. Trained on the OU-MVLP dataset (6,000 subjects) and evaluated on unseen CASIA-B across normal, bag-carrying, and coat-wearing conditions, our pipeline achieves 96.91\% single-view and 99.49\% three-view accuracy on OU-MVLP, and attains 100\% accuracy on CASIA-B with three views.  \nocite{*}
\end{abstract}


\begin{highlights}
\item Two-stage CNN-Transformer pipeline enables flexible multi-view gait fusion.
\item Cross-view curricula improve OU-MVLP average accuracy from 91.3\% to 93.5\%.
\item Fusing three OU-MVLP views reaches 99.49\% rank-1 identification accuracy.
\end{highlights}



\maketitle

\section{Introduction}\label{sec:intro}

Gait---the manner in which a person walks---is a behavioral biometric trait that can be captured unobtrusively at a distance, without requiring the subject's cooperation or awareness. This makes gait recognition suitable for many applications where face recognition is not viable. However, it faces a fundamental challenge: the same person's silhouette, and consequently their Gait Energy Image (\GEI{}) representation \citep{han2006individual}, changes substantially with the camera's viewing angle relative to the walking direction. In our experiments, the cosine similarity between GEIs of different subjects from the same view can be higher than that between GEIs of the same subject from different views, while humans can still visually distinguish them. We attribute this issue to the similarity metric itself. With limited low-cost options available, primarily Euclidean and cosine distances, we must transform GEIs into a representation where these standard metrics are more effective.

Two broad families of approaches have been proposed to address view variance: (1) \emph{view-invariant feature learning}, which attempts to learn a single representation robust to viewpoint changes, often through adversarial training, canonical pose transformation, or 3D model fitting; and (2) \emph{cross-view learning}, which explicitly trains discriminative models across pairs or sets of views using metric learning objectives. Our work follows the second family but introduces a two-stage decomposition motivated by limited computational resources. Rather than training a single deep network end-to-end across all views---which typically requires substantial GPU memory and training time---we first train a shallow \CNN{} under a Siamese/triplet-loss objective \citep{schroff2015facenet} to produce a per-view embedding, and subsequently train a sequence-oriented Transformer encoder \citep{vaswani2017attention} to refine and fuse these embeddings.

A conceptual contribution of this work is an architectural analogy between multi-view gait recognition and machine translation or natural language understanding, which we use to motivate---rather than to assert as an empirical property of gait data---the proposed two-stage decomposition. Under this analogy, each view-specific \GEI{} is loosely compared to a sentence written in a different "dialect" describing a person's identity: Stage 1 (the \CNN{}) acts as a translator, mapping each \GEI{} into a common embedding space, while Stage 2 (the Transformer) acts as a reader, taking one or more such embeddings and reasoning over them to produce a refined, more discriminative representation. We revisit this analogy in Section~\ref{sec:stage2}, clarifying that it serves solely to justify our architectural choice. Its practical benefit is that, since Stage 2 operates on a set of embeddings rather than a fixed number of raw images, the same trained model can flexibly handle one, two, or more input views at inference time, enabling natural multi-view fusion without retraining or architectural changes.

The remainder of this paper is organized as follows. Section~\ref{sec:related} reviews related work in gait recognition, metric learning, and Transformer-based sequence modeling. Section~\ref{sec:method} describes the proposed two-stage architecture and the curriculum-style cross-view training strategies for each stage. Section~\ref{sec:setup} details the experimental setup, datasets, and implementation specifics. Section~\ref{sec:results} presents and discusses results on OU-MVLP and cross-dataset evaluations on CASIA-B. Section~\ref{sec:conclusion} concludes the paper and outlines directions for future work.

\section{Related Work}\label{sec:related}

\subsection{Gait Representation and Gait Energy Images}

The \GEI{} averages silhouettes over a full gait cycle \citep{han2006individual}, giving a compact spatio-temporal representation. It has been widely used in appearance-based gait recognition because it is robust to segmentation noise, has a fixed size, and comes as a single image, which makes it easy to feed into standard image-based \CNN{} architectures. GEINet \citep{shiraga2016geinet} and similar convolutional models showed that \CNN{}s trained on \GEI{}s can work well across different views without needing manually designed features. More recently, set-based methods like GaitSet \citep{chao2019gaitset} treat gait sequences as unordered sets of silhouettes and combine \CNN{}s with set-pooling to achieve strong cross-view performance. Part-based approaches such as GaitPart \citep{fan2020gaitpart} go further by modeling different body regions with separate short-term temporal modules.

\subsection{Metric Learning: Siamese Networks and Triplet Loss}

Siamese networks trained with contrastive or triplet losses are a common choice for verification-style tasks. Their goal is to learn an embedding space where samples from the same identity are close to each other and samples from different identities are far apart. Triplet loss, which gained popularity through face recognition \citep{schroff2015facenet}, directly optimizes the relative distances between an anchor, a positive sample (same identity), and a negative sample (different identity). It has since been extended to other biometric domains, including gait. In our work, we use cosine distance instead of Euclidean distance as the similarity metric within the triplet loss. We found this to be a good fit because \CNN{} feature embeddings tend to be high-dimensional and direction-sensitive, which makes cosine distance a more natural choice.

\subsection{Transformer Encoders for Sequence Modeling}

The Transformer \citep{vaswani2017attention} was originally developed for machine translation. It replaces recurrence with self-attention, allowing the model to weigh the importance of different elements in a set or sequence when forming a contextualized representation. Although Transformers are typically used on sequences of words or image patches, self-attention can handle any set of feature vectors, including our per-view gait embeddings, as long as we provide positional or view-specific information about the input. In our approach, we add a standard sinusoidal positional encoding to the \CNN{}-derived gait embeddings, pass them through a 12-layer Transformer encoder, and then apply mean pooling over the output sequence. This yields a single refined embedding, much like how sentence-level representations are often obtained by pooling token-level Transformer outputs.

\subsection{Multi-View Gait Datasets}

The OU-ISIR Multi-View Large Population Dataset (OU-MVLP) \citep{takemura2018multi} is one of the largest publicly available multi-view gait datasets, providing \GEI{} or silhouette sequences for a very large number of subjects captured from up to 14 azimuth viewing angles, making it a standard large-scale benchmark for cross-view gait recognition research. CASIA-B \citep{yu2006framework} is a smaller but widely used benchmark that additionally varies walking condition (normal walking, walking with a bag, and walking while wearing a coat) across 11 viewing angles, and is frequently used to test the robustness of gait recognition systems to both viewpoint and appearance/clothing variation. In this work, OU-MVLP is used for both training and in-domain testing, while CASIA-B is reserved exclusively for cross-dataset, cross-condition generalization testing.

\section{Methodology}\label{sec:method}

\subsection{Problem Formulation}

Let $x_i^v$ denote the \GEI{} of subject $i$ captured at view angle $v \in V$, where $V$ is the set of available azimuth angles (e.g., $\{0^\circ, 15^\circ, 30^\circ, \dots, 270^\circ\}$, depending on the dataset). The goal is to learn a function $f$ that maps one or more \GEI{}s of a subject, potentially captured at different, unknown views, to an embedding $e_i \in \mathbb{R}^{128}$, such that for any two subjects $i \neq j$:
\[
\cos(e_i, e_i') > \cos(e_i, e_j)
\]
for embeddings $e_i, e_i'$ derived from different samples of subject $i$, and $e_j$ derived from a sample of a different subject $j$. We decompose $f$ into two learned sub-functions, $f = g \circ h$, described below.

\subsection{Stage 1: Siamese CNN Embedding with Triplet Loss}\label{sec:stage1}

\textbf{Architecture.} The first-stage network $h$ is a shallow CNN. It consists of three convolutional blocks, each with a $7\times7$ convolution, batch normalization \citep{ioffe2015batch}, a SiLU (Swish) activation \citep{elfwing2018sigmoid}, and $2\times2$ max pooling. After these blocks, a two-layer fully connected head (with dropout rate 0.5) flattens the convolutional features and reduces them to a 128-dimensional embedding. The channel widths of the three blocks are 16, 64, and 256, respectively. 

\textbf{Training objective.} We train the network using triplet loss with cosine distance:
\[
\mathcal{L}_{triplet} = \max\bigl(0,\; \cos(h(a), h(n)) - \cos(h(a), h(p)) + \alpha\bigr)
\]
Here, $a$, $p$, and $n$ are anchor, positive (same identity), and negative (different identity) \GEI{} samples, and $\alpha$ is the margin.

\textbf{Cross-view curriculum training.} Instead of training on all views at once, Stage 1 follows a curriculum with two phases.

\begin{figure}
  \centering
  \includegraphics[width=\linewidth]{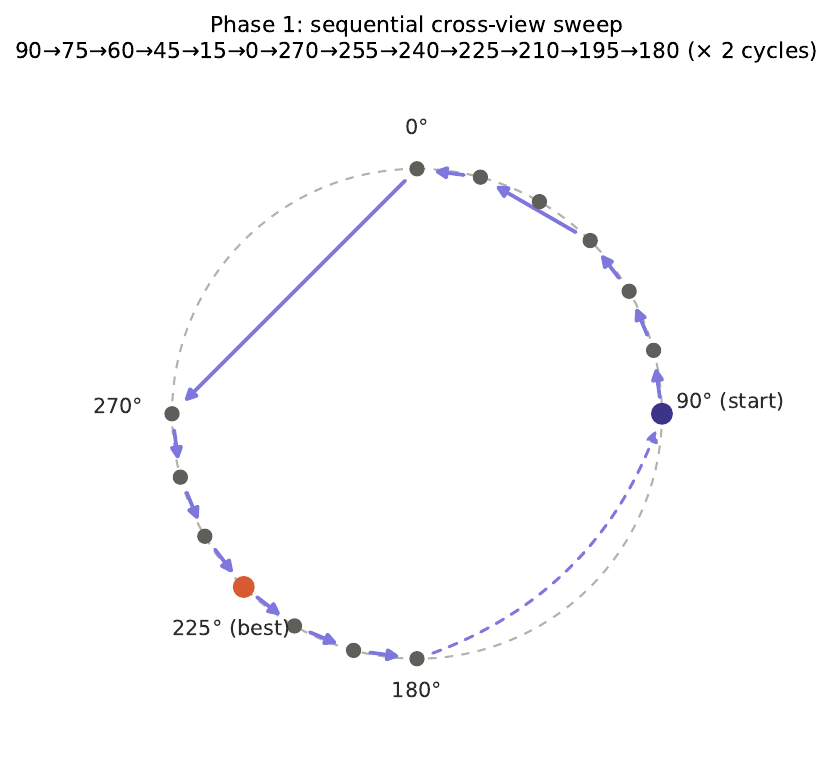}
  \caption{Phase 1 sequential cross-view sweep. Starting from 90$^\circ$ (dark marker), training moves through nearby views before crossing to the opposite side, then continues around to the best-performing 225$^\circ$ view (highlighted marker). The dashed arrow shows the wrap-around back to 90$^\circ$ for the second cycle.}
  \label{fig:phase1}
\end{figure}

\emph{Phase 1 (sequential sweep).} We start training at the 90$^\circ$ view. After we get the best validation checkpoint at 90$^\circ$, we move to 75$^\circ$, then continue through the remaining views in this cyclic order:
\begin{center}
\footnotesize
$90^\circ \rightarrow 75^\circ \rightarrow 60^\circ \rightarrow 45^\circ \rightarrow 15^\circ \rightarrow 0^\circ \rightarrow 270^\circ$ \\
$\rightarrow\, 255^\circ \rightarrow 240^\circ \rightarrow 225^\circ \rightarrow 210^\circ \rightarrow 195^\circ \rightarrow 180^\circ \rightarrow 90^\circ \rightarrow \dots$
\end{center}
We repeat this full cycle twice. Figure~\ref{fig:phase1} illustrates this sweep.

\begin{figure}
  \centering
  \includegraphics[width=\linewidth]{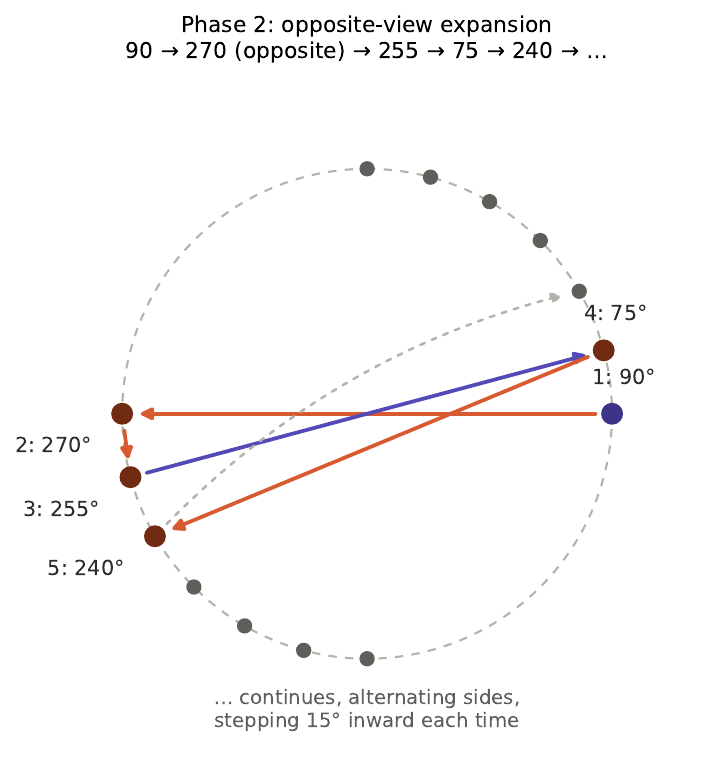}
  \caption{Phase 2 opposite-view expansion. Training alternates between the two poles (90$^\circ$ and 270$^\circ$), stepping 15$^\circ$ inward from each pole on alternating turns.}
  \label{fig:phase2}
\end{figure}

\emph{Phase 2 (opposite-view expansion).} After Phase 1, we switch to a different schedule. From the current view, we move to the opposite view (180$^\circ$ away), then to the view immediately to the right of that opposite view, alternating back and forth. For example, starting at 90$^\circ$, we next train on 270$^\circ$ (opposite), then 255$^\circ$ (adjacent to 270$^\circ$), then 75$^\circ$ (adjacent to 90$^\circ$ on the other side), then 240$^\circ$, and so on. This expands outward from the two poles of the view circle in a staggered way, as shown in Figure~\ref{fig:phase2}.

\textbf{Stage 1 outcome.} After both phases, the Stage 1 CNN alone reaches 96.43\% test accuracy at 225$^\circ$ (its best view) on OU-MVLP, but only 82.27\% at 0$^\circ$ (the most difficult view). The average accuracy across all views is about 91.3\%. This gap is what motivates Stage 2.

\subsection{Stage 2: Transformer Encoder Re-Embedding}\label{sec:stage2}

\textbf{Motivation.} The design of Stage 2 is loosely inspired by sequence modeling in natural language processing. We use this analogy purely as a way to guide our architectural choices, not because we believe gait embeddings have linguistic properties. In machine translation, sentences with very different surface forms (vocabulary, word order) can still convey the same meaning after being mapped to a shared embedding space. Transformer encoders are designed to reason over a set of such embeddings rather than a single fixed input. By analogy, Stage 1 acts like a translator—it maps view-specific \GEI{}s, which look very different at the pixel level, into embeddings that are meant to preserve identity information regardless of viewing angle. Stage 2 then acts like a reader that jointly reasons over one or more of these embeddings to infer the subject's identity. We use this analogy only to justify our choice of a permutation-tolerant, variable-length encoder for Stage 2. The practical outcome, which we evaluate directly in Section~\ref{sec:results}, is that a single trained Stage 2 model can accept and fuse any number of Stage-1 embeddings at inference time. This contrasts with earlier single-stage pipelines, which typically fix the number of input images at training time.

\textbf{Architecture.} Stage 2 uses a standard sinusoidal positional encoding followed by a 12-layer Transformer encoder with 8 attention heads, a feed-forward dimension of 512, and a dropout rate of 0.5. The model dimension is set to 128 to match the Stage 1 embedding size. Given a set of one or more Stage-1 embeddings for a subject, the Transformer processes them as a sequence. We then apply mean pooling over the output sequence to obtain a final 128-dimensional representation. We train Stage 2 using the same cosine-based triplet loss as Stage 1, but now it operates on the Transformer outputs rather than the raw CNN outputs.

\textbf{Progressive multi-view group curriculum.} For Stage 2, we train on groups of four views that are evenly spaced around the view circle. This encourages the model to learn how to combine information from widely separated views, which helps it generalize to views not included in the current training group. Training proceeds through a sequence of four-view groups. Each new group is shifted by 15$^\circ$ from the previous one (wrapping around the circle), and we initialize training from the best checkpoint of the prior group:
\[
\{225^\circ, 90^\circ, 45^\circ, 270^\circ\} \rightarrow \{210^\circ, 75^\circ, 30^\circ, 255^\circ\} \rightarrow \dots
\]
After this four-view curriculum, we do an additional round of fine-tuning using an extended six-view group:
\[
\{255^\circ, 210^\circ, 90^\circ, 75^\circ, 45^\circ, 270^\circ\}
\]
This six-view group gives us the best overall results in Stage 2 (see Section~\ref{sec:results}). 

\textbf{Multi-view fusion at inference.} Since Stage 2 works on a set of embeddings rather than a fixed-size input, we can pass any number of Stage-1 embeddings for the same subject into the trained Transformer at inference time. This includes combinations of views that were never seen together during training. The model produces a single fused embedding as output. This flexibility is a key advantage over traditional single-stage pipelines, which usually require retraining or architectural changes to handle a different number of input views. We test this fusion capability with groups of one, two, and three views in Section~\ref{sec:results}.

\section{Experimental Setup}\label{sec:setup}

\subsection{Datasets}

\textbf{OU-MVLP (training and in-domain test).} The OU-ISIR Gait Database, Multi-View Large Population Dataset \citep{takemura2018multi}, is used for both training and in-domain testing. The dataset is split at the subject level into 6,000 subjects for training, 2,000 subjects for validation, and 2,000 subjects for testing, with no subject overlap between splits. \GEI{}s from 14 azimuth viewing angles ($0^\circ$, $15^\circ$, $30^\circ$, $45^\circ$, $60^\circ$, $75^\circ$, $90^\circ$, $180^\circ$, $195^\circ$, $210^\circ$, $225^\circ$, $240^\circ$, $255^\circ$, $270^\circ$) are used.

\textbf{CASIA-B (cross-dataset test only).} CASIA-B \citep{yu2006framework} is used exclusively as an out-of-domain test set and is never used in any training or validation step. Three walking conditions are evaluated: \emph{nm} (normal walking, no carried objects), \emph{bg} (walking while carrying a bag), and \emph{cl} (walking while wearing a coat), at the $0^\circ$, $90^\circ$, and $180^\circ$ viewing angles.

\subsection{Implementation Details}

All models are trained on two NVIDIA T4 GPUs (15\,GB each) provided by the Kaggle platform, reflecting a deliberately constrained computational budget relative to much of the gait recognition literature, which often assumes access to higher-end accelerators. The Stage 1 \CNN{} produces 128-dimensional embeddings from single-channel \GEI{} input using three convolutional blocks (16 $\rightarrow$ 64 $\rightarrow$ 256 channels) with SiLU activations and batch normalization. The Stage 2 Transformer encoder has 12 layers, 8 attention heads, feed-forward dimension 512, and operates on the 128-dimensional Stage 1 embeddings. Both stages are trained with triplet loss using cosine distance as the similarity metric, with a dropout rate of 0.5 applied in the fully connected / feed-forward sublayers.

\subsection{Evaluation Protocol}\label{sec:protocol}

For each test view (or combination of views), we follow the standard identification protocol for gait recognition. We compare gallery and probe embeddings using cosine similarity, and report identification accuracy as the rank-1 match rate against the gallery. When evaluating multi-view fusion, we pass the embeddings from the specified set of views together into the Stage 2 Transformer encoder. This gives us a single fused embedding, which we then match against the gallery in the same way.

\section{Results and Discussion}\label{sec:results}

\subsection{Single-View Accuracy on OU-MVLP}

Table~\ref{tbl:singleview} reports rank-1 test accuracy at each of the 14 OU-MVLP views, comparing the Stage 1 \CNN{} embedding directly against the Stage 2 Transformer re-embedding (single view, i.e. $N=1$). The Stage 1 \CNN{} alone already achieves strong performance on side-facing and oblique views (e.g., $225^\circ$: 96.43\%), but noticeably weaker performance on the frontal ($0^\circ$: 82.10\%) and rear-facing ($180^\circ$: 88.90\%) views, which are known to carry less discriminative silhouette information for gait. Re-embedding the Stage 1 outputs with the Stage 2 Transformer encoder improves accuracy at nearly every view, including the previously weakest view ($0^\circ$: 82.10\% $\rightarrow$ 83.49\%), and raises the best single-view accuracy from 96.43\% to 96.91\% at $225^\circ$. The average single-view test accuracy across all 14 views rises to 93.52\%, up from approximately 91.3\% after Stage 1 alone.

\begin{table}
\caption{Single-view rank-1 accuracy (\%) on OU-MVLP: Stage 1 (\CNN) vs. Stage 2 (Transformer).}\label{tbl:singleview}
\begin{tabular*}{\tblwidth}{@{}LLL@{}}
\toprule
View & Stage 1 (\CNN) & Stage 2 (Transformer) \\
\midrule
$000^\circ$ & 82.10 & 83.49 \\
$015^\circ$ & 88.84 & 90.57 \\
$030^\circ$ & 93.62 & 95.04 \\
$045^\circ$ & 94.60 & 95.25 \\
$060^\circ$ & 93.70 & 94.61 \\
$075^\circ$ & 94.56 & 95.44 \\
$090^\circ$ & 94.45 & 95.39 \\
$180^\circ$ & 88.90 & 88.47 \\
$195^\circ$ & 93.74 & 94.45 \\
$210^\circ$ & 95.18 & 95.92 \\
$225^\circ$ & 96.43 & 96.91 \\
$240^\circ$ & 92.86 & 93.87 \\
$255^\circ$ & 94.16 & 94.30 \\
$270^\circ$ & 94.92 & 95.61 \\
\midrule
Average & $\sim$91.3 & 93.52 \\
\bottomrule
\end{tabular*}
\end{table}

\subsection{Multi-View Fusion on OU-MVLP}

\textbf{Two-view fusion.} 

\begin{table}[htbp]
\raggedright
\caption{Two-view fusion rank-1 accuracy (\%) on OU-MVLP.}
\label{tbl:twoview}
\begin{tabular}{c|rrrrr}
\hline
& 000 & 015 & 030 & 045 & 060 \\
\hline
000 & 83.54 & 97.32 & 98.03 & 98.34 & 97.16 \\
015 & 97.32 & 90.52 & 96.83 & 97.45 & 97.69 \\
030 & 98.09 & 96.83 & 95.04 & 96.63 & 97.18 \\
045 & 98.34 & 97.45 & 96.58 & 95.25 & 96.82 \\
060 & 97.16 & 97.69 & 97.18 & 96.82 & 94.61 \\
075 & 97.84 & 97.36 & 97.09 & 97.47 & 96.88 \\
090 & 97.20 & 97.33 & 97.63 & 97.82 & 97.00 \\
180 & 96.47 & 96.41 & 97.46 & 98.08 & 97.22 \\
195 & 96.44 & 97.80 & 98.72 & 99.02 & 98.17 \\
210 & 97.27 & 98.18 & 98.75 & 98.86 & 98.54 \\
225 & 98.26 & 98.31 & 98.73 & 99.18 & 98.80 \\
240 & 96.69 & 97.97 & 98.32 & 98.87 & 97.99 \\
255 & 96.15 & 97.97 & 98.32 & 98.49 & 98.29 \\
270 & 97.19 & 98.10 & 98.23 & 98.63 & 98.27 \\
\hline
\end{tabular}
\bigskip

\begin{tabular}{c|rrrrr}
\hline
& 075 & 090 & 180 & 195 & 210 \\
\hline
000 & 97.78 & 97.20 & 96.53 & 96.57 & 97.33 \\
015 & 97.41 & 97.33 & 96.41 & 97.80 & 98.18 \\
030 & 97.14 & 97.63 & 97.46 & 98.67 & 98.75 \\
045 & 97.47 & 97.82 & 98.08 & 99.02 & 98.86 \\
060 & 96.88 & 97.00 & 97.22 & 98.17 & 98.54 \\
075 & 95.44 & 96.96 & 97.21 & 98.73 & 98.67 \\
090 & 96.96 & 95.39 & 97.12 & 98.38 & 98.80 \\
180 & 96.93 & 97.17 & 88.52 & 98.20 & 98.35 \\
195 & 98.73 & 98.43 & 98.14 & 94.45 & 97.45 \\
210 & 98.67 & 98.85 & 98.18 & 97.56 & 95.92 \\
225 & 98.86 & 99.09 & 98.44 & 98.41 & 97.82 \\
240 & 98.32 & 98.22 & 97.79 & 97.71 & 97.81 \\
255 & 98.68 & 98.40 & 97.70 & 97.87 & 98.31 \\
270 & 98.59 & 98.59 & 97.92 & 98.30 & 98.21 \\
\hline
\end{tabular}
\bigskip

\begin{tabular}{c|rrrr}
\hline
& 225 & 240 & 255 & 270 \\
\hline
000 & 98.26 & 96.69 & 96.15 & 97.25 \\
015 & 98.31 & 97.97 & 97.97 & 98.10 \\
030 & 98.73 & 98.25 & 98.32 & 98.27 \\
045 & 99.18 & 98.87 & 98.49 & 98.63 \\
060 & 98.74 & 97.99 & 98.29 & 98.27 \\
075 & 98.91 & 98.32 & 98.59 & 98.59 \\
090 & 99.09 & 98.22 & 98.35 & 98.59 \\
180 & 98.44 & 97.79 & 97.64 & 97.92 \\
195 & 98.41 & 97.58 & 97.92 & 98.30 \\
210 & 97.82 & 97.81 & 98.31 & 98.21 \\
225 & 96.91 & 97.93 & 98.22 & 98.36 \\
240 & 97.93 & 93.87 & 96.89 & 97.41 \\
255 & 98.22 & 96.89 & 94.58 & 96.96 \\
270 & 98.36 & 97.47 & 96.91 & 95.61 \\
\hline
\end{tabular}
\end{table}

\begin{figure}
  \centering
  \includegraphics[width=\linewidth]{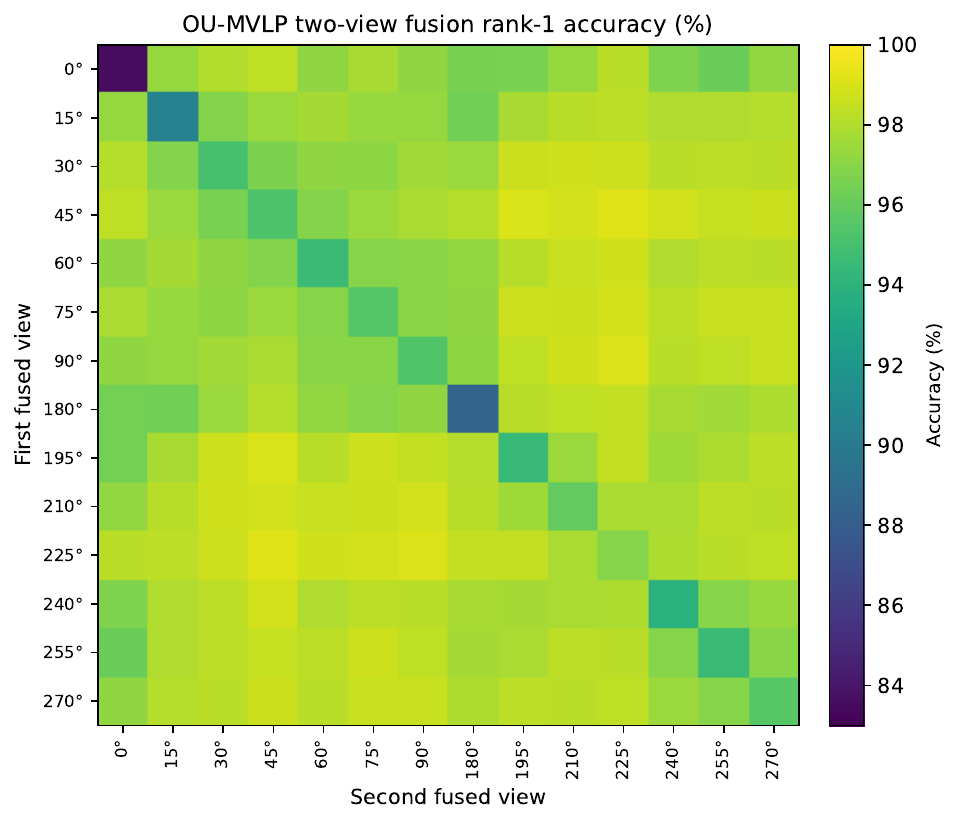}
  \caption{OU-MVLP two-view fusion rank-1 accuracy (\%). Diagonal cells are single-view accuracy (from Table~\ref{tbl:singleview}); off-diagonal cells show the accuracy obtained by fusing the corresponding pair of views through the Stage 2 Transformer encoder. Fusion consistently and substantially improves on single-view accuracy, especially for the weakest single views ($0^\circ$, $180^\circ$).}
  \label{fig:heatmap}
\end{figure}

Figure~\ref{fig:heatmap} and Table~\ref{tbl:twoview} visualize rank-1 accuracy when fusing embeddings from every pairwise combination of two views via the Stage 2 Transformer encoder (diagonal cells correspond to the single-view case from Table~\ref{tbl:singleview}). Fusing just two views produces a substantial and consistent accuracy improvement over single-view inference, including for the previously difficult $0^\circ$ and $180^\circ$ views: for example, fusing $0^\circ$ with $45^\circ$ raises accuracy to 98.34\%, compared to 83.49\% for $0^\circ$ or 95.25\% for $45^\circ$  alone.

\textbf{Three-view fusion.} Table~\ref{tbl:threeview} reports accuracy for representative three-view combinations spaced around the view circle. Three-view fusion further improves accuracy across the board, with the best combination ($045$-$195$-$225$) reaching 99.49\%, illustrating that the Stage 2 model benefits from additional, spatially diverse view information even beyond what two views can provide.

\begin{table}
\caption{Three-view fusion rank-1 accuracy (\%) on OU-MVLP.}\label{tbl:threeview}
\begin{tabular*}{\tblwidth}{@{}LL@{}}
\toprule
View combination & Accuracy \\
\midrule
000-090-180 & 99.00 \\
045-195-225 & 99.49 \\
210-060-030 & 99.07 \\
255-075-015 & 98.98 \\
270-180-090 & 99.15 \\
\bottomrule
\end{tabular*}
\end{table}

\subsection{Cross-Dataset Generalization on CASIA-B}

CASIA-B is used solely as a held-out, cross-dataset test set to assess generalization beyond OU-MVLP, including under bag-carrying and coat-wearing conditions not represented during training. Table~\ref{tbl:casiab} reports single-view accuracy for the Stage 1 \CNN{} and the Stage 2 Transformer across all three walking conditions and the $0^\circ$/$90^\circ$/$180^\circ$ views. Despite never being trained on CASIA-B, the Stage 2 model generalizes strongly: three-view fusion "Table~\ref{tbl:casiab_three}" ($0^\circ$-$90^\circ$-$180^\circ$) reaches 100\% rank-1 accuracy for all three walking conditions (\emph{nm}, \emph{bg}, \emph{cl}). Notably, Stage 2 improves substantially over Stage 1 alone at $0^\circ$ under both the \emph{nm} condition (80.64\% $\rightarrow$ 95.16\%) and the \emph{bg} condition (87.90\% $\rightarrow$ 95.96\%), reinforcing the pattern already observed on OU-MVLP: the Transformer re-embedding stage disproportionately benefits the most viewpoint-degraded conditions.

\begin{table}
\caption{CASIA-B single-view rank-1 accuracy (\%): Stage 1 (\CNN) vs. Stage 2 (Transformer).}\label{tbl:casiab}
\begin{tabular*}{\tblwidth}{@{}LLLLL@{}}
\toprule
Condition & View & Stage 1 & Stage 2 \\
\midrule
nm & $0^\circ$   & 80.64 & 95.16 \\
nm & $90^\circ$  & 83.87 & 85.48 \\
nm & $180^\circ$ & 83.06 & 91.93 \\
bg & $0^\circ$   & 87.90 & 95.96 \\
bg & $90^\circ$  & 89.51 & 88.70 \\
bg & $180^\circ$ & 87.09 & 93.54 \\
cl & $0^\circ$   & 97.58 & 100.00 \\
cl & $90^\circ$  & 100.00 & 99.19 \\
cl & $180^\circ$ & 97.58 & 96.77 \\
\bottomrule
\end{tabular*}
\end{table}

\begin{table}
\caption{CASIA-B three-view fusion rank-1 accuracy (\%):}\label{tbl:casiab_three}
\begin{tabular*}{\tblwidth}{@{}LLLL@{}}
\toprule
Condition & View combination & Accuracy \\
\midrule
nm & $0^\circ-90^\circ-180^\circ$   & 100.00 \\
bg & $0^\circ-90^\circ-180^\circ$   & 100.00 \\
cl & $0^\circ-90^\circ-180^\circ$   & 100.00 \\
\bottomrule
\end{tabular*}
\end{table}

\subsection{Discussion}

\textbf{Persistent difficulty of the frontal/rear views.} Across both datasets, the $0^\circ$ (and to a lesser extent $180^\circ$) view remains the hardest single-view condition, consistent with the general finding in the gait recognition literature that near-frontal and near-rear views convey less discriminative silhouette shape information than side views. Multi-view fusion is particularly effective at compensating for this weakness, since even a single additional, more informative view (e.g., a side view) sharply raises accuracy when fused with a frontal embedding.

\textbf{Computational efficiency as a design constraint.} A central motivation of this work is that competitive multi-view gait recognition accuracy can be obtained without large-scale compute. Both stages use comparatively shallow architectures -- a 3-block \CNN{} and a 128-dimensional Transformer encoder. Because Stage 1 embeddings are only 128-dimensional and Stage 2 operates on these compact embeddings rather than raw images, inference is inexpensive to run and cheap to store.

\textbf{Flexibility of the two-stage design.} Unlike prior approaches that fix the number of input images (typically one or two) at both training and inference time, treating Stage 1 embeddings as elements of a set that Stage 2 can consume in variable quantity allows the same trained Stage 2 model to be used with one, two, three, or more views at inference, without retraining. This is reflected in the monotonic accuracy improvements observed above as additional views are fused.

\section{Conclusion}\label{sec:conclusion}

This paper presented a two-stage framework for multi-view gait recognition that is computationally efficient. Stage 1 uses a shallow Siamese CNN trained with cosine-based triplet loss to produce per-view embeddings. Stage 2 uses a 12-layer Transformer encoder to re-embed and fuse one or more of these embeddings. We drew an architectural analogy to translation and sentence-level reasoning in natural language processing (Section~\ref{sec:stage2}) to motivate this design, but we do not claim that gait data has any linguistic structure. The main practical benefit is that the framework naturally supports flexible multi-view fusion with a variable number of views at inference time.

On OU-MVLP, our approach improves single-view test accuracy from 96.43\% (Stage 1 only) to 96.91\% (Stage 2) at the best-performing view. The cross-view average increases from about 91.3\% to 93.52\%, and three-view fusion reaches 99.49\% accuracy. On CASIA-B, which was never used during training, three-view fusion achieves 100\% accuracy across normal walking, bag-carrying, and coat-wearing conditions.

\printcredits

\bibliographystyle{cas-model2-names}

\bibliography{cas-refs}

@article{han2006individual,
  title   = {Individual recognition using gait energy image},
  author  = {Han, Ju and Bhanu, Bir},
  journal = {IEEE Transactions on Pattern Analysis and Machine Intelligence},
  volume  = {28},
  number  = {2},
  pages   = {316--322},
  year    = {2006}
}

@inproceedings{yu2006framework,
  title     = {A framework for evaluating the effect of view angle, clothing and carrying condition on gait recognition},
  author    = {Yu, Shiqi and Tan, Daoliang and Tan, Tieniu},
  booktitle = {18th International Conference on Pattern Recognition (ICPR)},
  pages     = {441--444},
  year      = {2006}
}

@inproceedings{shiraga2016geinet,
  title     = {GEINet: View-invariant gait recognition using a convolutional neural network},
  author    = {Shiraga, Kohei and Makihara, Yasushi and Muramatsu, Daigo and Echigo, Tomio and Yagi, Yasushi},
  booktitle = {International Conference on Biometrics (ICB)},
  pages     = {1--8},
  year      = {2016}
}

@article{takemura2018multi,
  title   = {Multi-view large population gait dataset and its performance evaluation for cross-view gait recognition},
  author  = {Takemura, Noriko and Makihara, Yasushi and Muramatsu, Daigo and Echigo, Tomio and Yagi, Yasushi},
  journal = {IPSJ Transactions on Computer Vision and Applications},
  volume  = {10},
  number  = {1},
  pages   = {4},
  year    = {2018}
}

@inproceedings{schroff2015facenet,
  title     = {FaceNet: A unified embedding for face recognition and clustering},
  author    = {Schroff, Florian and Kalenichenko, Dmitry and Philbin, James},
  booktitle = {IEEE Conference on Computer Vision and Pattern Recognition (CVPR)},
  pages     = {815--823},
  year      = {2015}
}

@inproceedings{vaswani2017attention,
  title     = {Attention is all you need},
  author    = {Vaswani, Ashish and Shazeer, Noam and Parmar, Niki and Uszkoreit, Jakob and Jones, Llion and Gomez, Aidan N and Kaiser, {\L}ukasz and Polosukhin, Illia},
  booktitle = {Advances in Neural Information Processing Systems (NeurIPS)},
  pages     = {5998--6008},
  year      = {2017}
}

@inproceedings{chao2019gaitset,
  title     = {GaitSet: Regarding gait as a set for cross-view gait recognition},
  author    = {Chao, Hanqing and He, Yiwei and Zhang, Junping and Feng, Jianfeng},
  booktitle = {AAAI Conference on Artificial Intelligence},
  volume    = {33},
  pages     = {8126--8133},
  year      = {2019}
}

@inproceedings{fan2020gaitpart,
  title     = {GaitPart: Temporal part-based model for gait recognition},
  author    = {Fan, Chao and Peng, Yunjie and Cao, Chunshui and Liu, Xu and Hou, Saihui and Chi, Chunfeng and Huang, Yongzhen and Li, Qing and He, Zhiqiang},
  booktitle = {IEEE Conference on Computer Vision and Pattern Recognition (CVPR)},
  pages     = {14225--14233},
  year      = {2020}
}

@article{elfwing2018sigmoid,
  title   = {Sigmoid-weighted linear units for neural network function approximation in reinforcement learning},
  author  = {Elfwing, Stefan and Uchibe, Eiji and Doya, Kenji},
  journal = {Neural Networks},
  volume  = {107},
  pages   = {3--11},
  year    = {2018}
}

@inproceedings{ioffe2015batch,
  title     = {Batch normalization: Accelerating deep network training by reducing internal covariate shift},
  author    = {Ioffe, Sergey and Szegedy, Christian},
  booktitle = {International Conference on Machine Learning (ICML)},
  pages     = {448--456},
  year      = {2015}
}



\end{document}